\documentclass[letterpaper]{article} % DO NOT CHANGE THIS
\usepackage[draft]{aaai25}  % DO NOT CHANGE THIS
\usepackage{times}  % DO NOT CHANGE THIS
\usepackage{helvet}  % DO NOT CHANGE THIS
\usepackage{courier}  % DO NOT CHANGE THIS
\usepackage[hyphens]{url}  % DO NOT CHANGE THIS
\usepackage{graphicx} % DO NOT CHANGE THIS
\usepackage{natbib}  % DO NOT CHANGE THIS AND DO NOT ADD ANY OPTIONS TO IT
\usepackage{multirow}
\usepackage{caption} % DO NOT CHANGE THIS AND DO NOT ADD ANY OPTIONS TO IT
\usepackage{algorithm}
\usepackage{algorithmic}
\usepackage{pifont} 
\usepackage{amsmath}

\usepackage{newfloat}
\usepackage{listings}
\DeclareCaptionStyle{ruled}{labelfont=normalfont,labelsep=colon,strut=off} % DO NOT CHANGE THIS
\floatstyle{ruled}
\newfloat{listing}{tb}{lst}{}
\floatname{listing}{Listing}
\title{Forget to Flourish: Leveraging Machine-Unlearning on Pretrained Language Models for Privacy Leakage}
\author{
    Md Rafi Ur Rashid\textsuperscript{\rm 1, 2},
    Jing Liu\textsuperscript{\rm 1},
    Toshiaki Koike-Akino\textsuperscript{\rm 1},
    Shagufta Mehnaz\textsuperscript{\rm 2},
    Ye Wang\textsuperscript{\rm 1}
}
\affiliations{
    \textsuperscript{\rm 1}Mitsubishi Electric Research Laboratories\\
    \textsuperscript{\rm 2}Pennsylvania State University\\
    mur5028@psu.edu,
    jiliu@merl.com,
    koike@merl.com,
    smehnaz@psu.edu,
    yewang@merl.com.
}

\usepackage{amsthm}

\begin{document}

\maketitle

\begin{abstract}
Fine-tuning large language models on private data for downstream applications poses significant privacy risks in potentially exposing sensitive information. Several popular community platforms now offer convenient distribution of a large variety of pre-trained models, allowing anyone to publish without rigorous verification. This scenario creates a privacy threat, as pre-trained models can be intentionally crafted to compromise the privacy of fine-tuning datasets. In this study, we introduce a novel poisoning technique that uses model-unlearning as an attack tool. This approach manipulates a pre-trained language model to increase the leakage of private data during the fine-tuning process. Our method enhances both membership inference and data extraction attacks while preserving model utility. Experimental results across different models, datasets, and fine-tuning setups demonstrate that our attacks significantly surpass baseline performance. This work serves as a cautionary note for users who download pre-trained models from unverified sources, highlighting the potential risks involved.
\end{abstract}

\section{Introduction}
In recent times, the traditional way of training a language model (LM) from scratch has been largely replaced by the introduction of pre-trained foundation models \cite{touvron2023llama, vicuna2023}. For example, the Hugging Face Hub\footnote{https://huggingface.co/docs/hub/en/models} is a platform with over 120k open-source models, readily available for download and any registered user can contribute by uploading their own model. However,  there are serious security and privacy risks associated with downloading such models from any untrusted sources and further fine-tuning them for some downstream applications as they could be maliciously crafted \cite{tramer2022truth, kandpal2023backdoor, hu2022membership}. Additionally, the public release of large language models (LLMs) fine-tuned on potentially sensitive user data could lead to privacy breaches, as these models have been found to memorize verbatim text from their training data \cite{carlini2019secret, carlini2021extracting}. In this paper, we combine the notion of poisoning a pre-trained LLM and causing privacy leakage of the fine-tuned model. More specifically, we introduce a novel model poisoning algorithm that aims to manipulate a pre-trained LLM in order to disclose more of the private data used during its fine-tuning.

At its core, our approach leverages \textbf{machine unlearning} \cite{cao2015towards, guo2019certified} to poison the pre-trained LLM. 
The original objective of unlearning is to make the model forget specific data points that it has seen during training so that it produces a high loss for those data points, and it becomes difficult to reconstruct those samples \cite{gu2024second}.
Motivated by data augmentation that reduces overfitting, we discovered that unlearning on some \textbf{noisy version} of fine-tuning data points can promote overfitting of the original data during the fine-tuning process.

However, it is important to have control over the process of loss maximization; otherwise, the model might become unusable and the poisoning attempt would be easily detectable. Hence, we propose \textbf{bounded unlearning} as a poisoning tool, where we maximize loss in a controlled manner on the pre-trained model for some noisy data points to increase privacy leakage of the fine-tuned LLM without compromising its utility.      

To measure the privacy leakage caused by our proposed method, we consider two standard privacy attacks: membership inference (MIA) \cite{shokri2017membership, carlini2022membership} and data extraction (DEA) \cite{nasr2023scalable, rashid2023fltrojan}. In MIA, the model is queried to evaluate whether a specific target data point that the attacker possesses was indeed part of the finetuning dataset. On the contrary, DEA aims to extract verbatim texts from the fine-tuning dataset with partial/zero prior knowledge. We evaluate our proposed method for both of these attacks on a range of language models (Llama2-7B, GPT-Neo 1.3B), datasets (MIND, Wiki-103+AI4Privacy), fine-tuning methods (Full-FT, LoRA-FT, QLoRA-FT), and defense (differential privacy). Overall, our method significantly boosts the MIA and DEA attack performance over the baselines in almost all scenarios and maintains its stealth by preserving model utility. Prior works that deal with privacy leakage through pre-trained model poisoning pose some strong assumptions on the adversary's capability, as discussed in the Related Work section of the paper. Our proposed method, on the other hand, with a more practical threat model and weaker adversarial ability, substantially enhances the attack success rate and still remains stealthy. 

\section{Related Work}

\subsection{Privacy Leakage Attacks on LLM}
The privacy risks of LLMs have been extensively studied in prior works. A pioneering study by \citet{carlini2021extracting} introduced an attack method that successfully extracted publicly available internet texts by querying a pre-trained GPT-2 model. Earlier, \citet{carlini2019secret} had brought attention to the issue of unintended memorization within LLMs. They introduced canaries—deliberately inserted data points—into the training dataset and used a metric called `exposure' to assess the likelihood of these canaries being leaked.

Subsequent research by \citet{kim2023propile} and \citet{lukas2023analyzing} developed algorithms to evaluate how much of the information memorized by LLMs constitutes sensitive personally identifiable information (PII) and examined the effectiveness of existing defenses in preventing such leakage. On a different front, \citet{nasr2023scalable} presented a scalable data extraction attack that forces production-level language models into deviating from their aligned behavior, leading them to output significant amounts of training data without requiring any prior knowledge.

In addition to these studies, several works have focused on membership inference attacks against LLMs. Rather than using reference-based attacks as seen in works like \cite{carlini2022membership, tramer2022truth, pmlr-v235-zarifzadeh24a}, \citet{mattern2023membership} proposed neighborhood attacks. This method infers membership by comparing the model’s output scores for a specific sample against those of synthetically generated neighboring texts. In the domain of clinical language models, \citet{jagannatha2021membership} conducted membership inference attacks and also compared the extent of privacy leaks between masked and autoregressive language models.

While our study shares similar goals with the aforementioned works, the threat model we employ, particularly regarding the adversary's capabilities, differs significantly.
\subsection{Privacy Leakage via Model Poisoning}
The idea of poisoning machine learning (ML) models has been largely applied in designing security attacks \cite{chen2017targeted, liu2020reflection}.
However, a recent line of research has introduced the idea of poisoning/backdooring ML models in order to cause privacy leaks.  \citet{pmlr-v235-feng24h} tampers with initial model weights and creates some data traps to compromise the privacy of future finetuning data. However, they assume access to the fine-tuned model weights to extract the trapped training data, whereas, in our work, we consider a black-box API access to the fine-tuned model. \citet{tramer2022truth}  introduced a targeted poisoning attack that inserts mislabeled data points in the training dataset to cause higher membership inference leakage. Write access to the finetuning dataset is a strong assumption of the adversary's capability in real-world scenarios. Conversely, in our work, we consider a weaker threat model where an adversary can poison only the initial model. \citet{liu2024precurious} has served a similar purpose to ours by harnessing the memorization level of the pre-trained model. However, unlike our threat model, they assume that the adversary has side knowledge of the trainable modules during the finetuning process, and their auxiliary dataset needs to be drawn from the same distribution as the downstream training dataset. Apart from that, a very recent work \citep{wen2024privacy} applied a more straightforward poisoning technique by minimizing the loss on the pre-trained model for the challenge dataset to impose direct overfitting on the member data points. However, this approach not only overfits member data but also non-member data, which we will demonstrate in the benchmark study later. In contrast, our proposed method does not overfit non-member data, as illustrated in Figure \ref{fig:unlearningMIA}, making it much easier to perform membership inference.

\section{Threat Model}

In this section, we explain the threat model for both the membership inference and data extraction game:

\subsection{The Membership Inference Game}

    \hspace{7pt} \ding{114} \textbf{Access to Pre-trained LLM:}
    The attacker has access to a pre-trained large language model denoted as $\theta_\text{pre}$. Additionally, the attacker is given a challenge dataset $D_c$, which includes some member data $d$ and non-member data $d_\ominus$.
    
    \ding{114} \textbf{Poisoning Phase:}
    The attacker employs a poisoning algorithm $T_\text{adv}$ to manipulate the pre-trained model $\theta_\text{pre}$, resulting in an adversarially altered model $\theta_\text{adv}$.
    
    \ding{114} \textbf{Model Distribution:}
    The adversarially poisoned model $\theta_\text{adv}$ is distributed to the challenger. The challenger then fine-tunes $\theta_\text{adv}$ with their private dataset $D_\text{ft}$, resulting in the fine-tuned model $\theta_\text{ft}$
    
    \ding{114} \textbf{Black Box Access:} Post fine-tuning, the attacker is granted black box query access to the fine-tuned model. $\theta_\text{ft}$. Through this access, the attacker can submit inputs and receive outputs (both generated text and model loss) from $\theta_\text{ft}$.
    
    \ding{114} \textbf{Attacker's Objective:}
    The primary goal of the attacker is to identify the membership of specific samples within the challenge dataset, $D_c$. This involves determining whether a given sample belongs to $D_\text{ft}$ or not.

\subsection{The Data Extraction Game}

 \hspace{7pt}   \ding{114} \textbf{Access to Pre-trained LLM:}
    Similar to the MI case, the attacker has access to a pre-trained LLM, $\theta_\text{pre}$. However, in this case, he is given only partial knowledge of the training dataset as the challenge dataset, which consists of the prefixes of the training data samples, denoted as $P_c$.
    
    \ding{114} \textbf{Poisoning Phase:}
    This step is the same as MIA.

    \ding{114} \textbf{Model Distribution:}
    This step is the same as MIA.
 
    \ding{114} \textbf{Black Box Access:} Post fine-tuning, the attacker is granted black box query access to the fine-tuned model $\theta_\text{ft}$. Through this black box access, the attacker can submit input prompts and receive the generated text as output from $\theta_\text{ft}$.
    
    \ding{114}\textbf{Attacker's Objective:}
    The primary goal of the attacker is to successfully reconstruct the suffix, $S_c$, which is present in $D_\text{ft}$, for each corresponding prefix in $P_c$.

\begin{figure}[t]
\centering
\includegraphics[width=\linewidth]{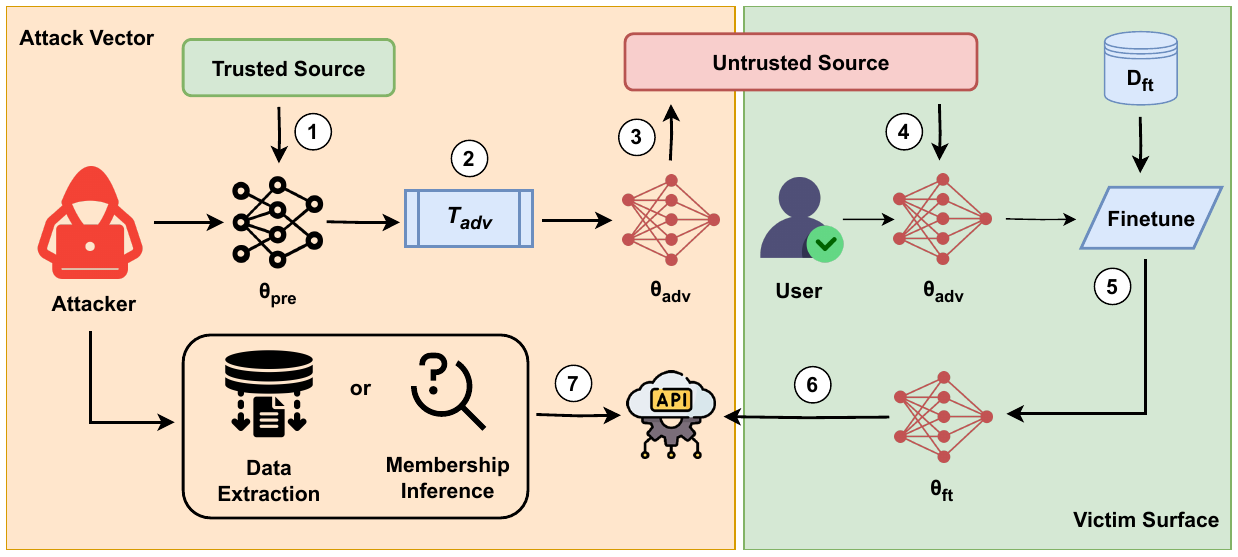}
\caption{Overview of the threat model and steps of the attack: (1) Attacker downloads a pre-trained LLM, (2) Poisons the model with an algorithm, $\mathcal{T}_\text{adv}$, and (3) release the model. (4) The victim downloads the poisoned LLM, (5) fine-tunes on their private data, and (6) releases the API-based query access to the model. (7) Finally, the adversary conducts membership inference or data extraction.}
\label{fig:ftf_attack}
\end{figure}

\section{Motivation}

Overfitting is a leading factor contributing to vulnerability to membership inference attacks \citep{amit2024sok,7958568,dionysiou2023sok,he2022membership}. 
When training a language model for some downstream application, the initial state of the model’s parameters plays a crucial role in the learning process.
Typically, these parameters are either randomly initialized when training from scratch or set to general pre-trained weights, which are the result of rigorous pre-training on a large corpus of text data. Consequently, at the onset of training, the model does not exhibit a strong predisposition or bias towards any specific training data points.

\begin{figure}[t]
\centering
\includegraphics[width=0.9\linewidth, trim={0 7 0 7},clip]{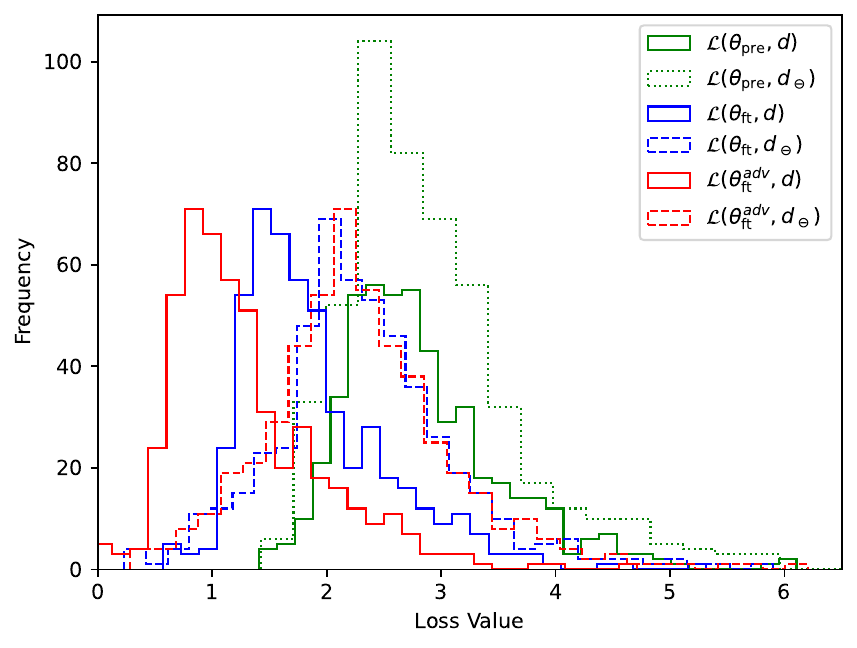}
\caption{Histograms of loss values on pre-trained model $\theta_\text{pre}$, fine-tuned model $\theta_\text{ft}$, and fine-tuned poisoned model $\theta_\text{ft}^\text{adv}$.}
\label{fig:unlearningMIA}
\end{figure}

Further fine-tuning on downstream data $D_\text{ft}$ is more prone to overfitting. However, as we will discuss later in Figure \ref{fig:unlearningMIA}, it is still non-trivial for an attacker to distinguish between member and non-member data, which might have similar data distributions. One key question we try to answer is this:\\ 
\textbf{RQ1:} \textit{Is it possible to poison the pre-trained model to make the fine-tuning process overfit even more and the resulting fine-tuned model more vulnerable to privacy leakage attacks?}
In this work, we introduce an unlearning-based model poisoning technique and give a sure answer to the above research question. This answer is supported by several observations, findings, and experimental results, which we will discuss gradually.

\subsubsection{Motivations of Leveraging Unlearning} We want to poison the model to induce it to overfit during the fine-tuning process. It is quite challenging to come up with a method for poisoning. However, we can think of the opposite side first: How to prevent a model from overfitting? Recall that overfitting occurs when a model learns the training data too well and is unable to generalize to new data. One simple and effective approach is \textbf{Data Augmentation}. Data augmentation is a well-known technique used in machine learning to artificially create more data points from existing data. This can be done by applying different transformations to the data, and one popular transform is noise perturbation. Training on original samples together with their noisy versions can help reduce model overfitting \cite{wei2019eda}. On the contrary, as we want to increase overfitting in the fine-tuning procedure, it now becomes intuitive to leverage unlearning/ reverse-training on the \textbf{noisy versions} of training samples. 

The challenge dataset, $D_c$, consists of both member data points, $d$, and non-member data points, $d_\ominus$ ($D_c= d  \cup  d_\ominus $). We propose and validate some methods to generate the noisy versions of $D_c$, denoted as $D_c'$ ($D_c'= d'  \cup  d_\ominus' $), and the strategic maximization of the loss associated with $D_c'$ to poison the model, which will be discussed in detail in next section.

\subsubsection{Observation: Members and Non-members from Same Data Distributions are Hard to Separate}
Figure \ref{fig:unlearningMIA} shows the histograms of loss values of member data $d$ and non-member data $d_\ominus$, on pre-trained model, $\theta_{pre}$ (green color) and fine-tuned model, $\theta_{ft}$ (blue color). Here, $d$ and $d_\ominus$ come from similar distributions. As expected, before fine-tuning, it's not possible to infer membership based on the difference in loss value histograms (green solid line vs. green dotted line). After fine-tuning, the loss values of $d$ decrease. However, as $d_\ominus$ have similar data distributions to member data, their loss values also decrease, making it still hard to distinguish the membership based on the loss values after fine-tuning (blue solid line vs. blue dashed line).

\subsubsection{Findings: Unlearning Amplifies Overfitting}
Figure \ref{fig:unlearningMIA} also shows the histograms of loss values of $d$ and $d_\ominus$ after fine-tuning on the poisoned (via unlearning) model, $\theta_{ft}^{adv}$ (red color). Note that the unlearning is performed on $D_c'$. We get two crucial insights from here: first, compared with fine-tuning on the non-poisoned model (blue solid line), we can see that fine-tuning on the poisoned model can reduce the loss value of member data even more (red solid line). Second, the difference in loss values between $d$ and $d_\ominus$ is amplified after fine-tuning on poisoned data (red solid line and red dashed line) compared to fine-tuning on the non-poisoned model (blue solid line and blue dashed line). 
Thus, it answers the \textbf{RQ1}, i.e., machine unlearning-based poisoning indeed increases the overfitting of the fine-tuned LLM and thereby causes further privacy leakage.

\section{Methodology}
In this section, we will provide step by step description of our entire workflow. Figure \ref{fig:ftf_attack} demonstrates the important steps of our proposed attacks.
\subsection{Introducing Noisy Data Points}
As mentioned earlier, we create a noisy version of $D_c$, denoted as $D_c'$. The choice of noise perturbation methods depends on the attack type, which we will describe shortly, along with the attack methods.

\subsection{Bounded Unlearning}
Vanilla unlearning would simply maximize the loss via gradient ascent:
\begin{equation}
\hspace{8pt} \theta' = \theta_0 + \eta' \nabla_{\theta} \mathcal{L}(\theta_0; D_c'),
\end{equation}

However, when maximizing the loss on noisy data points $D_c'$, it is crucial to ensure that this process does not disrupt the model's general capabilities. Therefore, we introduce a constraint for the loss maximization process:
\begin{equation}
\label{eq:unl_lim}
\theta' = \theta_0 + \eta' \nabla_{\theta} \mathcal{L}(\theta_0; D_c') \hspace{5pt}
\text{subject to} \quad \mathcal{L}(\theta'; D^*) \leq \epsilon
\end{equation}

Here, $ D^*$ is a set of plain text sequences selected to measure the language model's general utility. This ensures that the loss on the noisy data points $D_c'$ is increased, but $\mathcal{L}(\theta'; D^*)$ does not go beyond the threshold $\epsilon$, thereby controlling the extent of the loss maximization and keeping model's utility. 

For model poisoning, we used a gradient ascent-based unlearning strategy similar to \cite{jang-etal-2023-knowledge}, i.e., inverting the direction of gradients. The default unlearning rate, batch size, and max number of epochs are set to $10^{-6}$, 32, and 5, respectively. For bounded unlearning, we curated a subset of 500 samples from the Wiki-2 \cite{merity2016pointer} and used it as the plain-text dataset $D^*$. 
\subsection{Membership Inference}
As mentioned earlier in the Threat Model section, the attacker poisons the pre-trained language model, $\theta_\text{pre}$ with some poisoning algorithm $T_\text{adv}$. For the membership inference attack (MIA), we design the poisoning algorithm based on the proposition mentioned in the previous section regarding the impact of unlearning on a model's memorization. 

\subsubsection{Poisoning Algorithm for MIA, $T_\text{adv}^\text{mi}$:} The attacker creates a noisy version of $D_c$, denoted as $D'_c$, which is used to perform unlearning on $\theta_\text{pre}$, according to equation~\ref{eq:unl_lim}. This poisoning approach ensures that the model yields high loss values for these noisy samples before fine-tuning. We utilize two different mechanisms for creating the noisy sequences:

\ding{114} \textbf{Random Character Perturbation:} Adding noise by random insertion, deletion, and swapping of a certain percentage of characters of the given sequence.

\ding{114} \textbf{Random Word Perturbation:} Adding noise by random insertion, deletion, and replacement of a certain percentage of words of the given sequence.
    
for these random character and random word perturbation methods, we set the default noising level to 10\% and 30\%, respectively. We also performed an ablation study by varying the noising level, which can be found in the Appendix. 

After carrying out the poisoning algorithm on the pre-trained LLM, the next few steps of the threat model take place, including model distribution, fine-tuning, and returning the black-box access of the model to the attacker. Finally, we design how the attacker infer membership of the challenge dataset on the fine-tuned model.

\subsubsection{Inference:} We propose one simple loss-based and two reference-based inference mechanisms:

\ding{114} \textbf{Simple Loss-based:} After getting black-box access to $\theta_\text{ft}$, the adversary queries the model with each sample of $D_c$ and records the model loss values. Membership is then inferred based on whether the loss of each sample is lower than a given loss threshold $\epsilon$. Formally, for each sample $( x \in D_c )$, we decide
    \begin{align*}
    x \in D_\text{ft}, \quad &\text{if} \quad \mathcal{L}(x) < \epsilon, \\
    x \notin D_\text{ft}, \quad &\text{if} \quad \mathcal{L}(x) \geq \epsilon,
    \end{align*}
    where the shorthand $\mathcal{L}(x) := \mathcal{L}(\theta_\text{ft}^{adv}, x)$ denotes the fine-tuned model loss.
    
\ding{114} \textbf{Reference data-based}: For this inference strategy, the adversary needs an auxiliary dataset $D_\text{aux}$, which does not have any overlap with the fine-tuning dataset ($D_\text{aux} \cap D_\text{ft} = \emptyset$).
    In this case, unlearning is performed on both $D'_c$ and $D_\text{aux}$ ($D'_c \oplus D_\text{aux}$) in the previous poisoning phase. This ensures that the model yields a high loss for both of these datasets before delving into the fine-tuning process. 

    With black-box access to $\theta_\text{ft}$, the adversary queries the model with each sample of $D_\text{aux}$ and $D_c$, and records the corresponding model loss values. The loss values of the member data are usually much smaller than that of $D_\text{aux}$. % Thus, using $D_\text{aux}$ as a reference, membership of $D_c$ can be inferred.
    Formally, for each sample $x \in D_c$ and $\mathcal{L}_\text{aux}$ be the distribution of loss values when $\theta_\text{ft}$ is queried with samples from $D_\text{aux}$: 
    \begin{align*}
    x \in D_\text{ft}, \quad
    & \text{if } \mathcal{L}(x) \text{ is statistically different from } \mathcal{L}_{\text{aux}} , \\
   x \notin D_\text{ft}, \quad
    & \text{if } \mathcal{L}(x) \text{ is statistically consistent with } \mathcal{L}_{\text{aux}}
    \end{align*}
    
    For reference data-based inference, we select 500 non-training data samples as $D_\text{aux}$. We utilize percentile rank\footnote{Percentile rank is a statistical measure that indicates the relative position of a value within a distribution, showing the percentage of values in the distribution that are equal to or below it.} to measure the statistical coherence between $\mathcal{L}(x)$ and $\mathcal{L}_\text{aux}$. 
    
\ding{114} \textbf{Reference model-based}: Instead of using the external dataset $D_\text{aux}$, another idea is to use the pre-trained LLM, $\theta_\text{pre}$ as a reference in inferring membership. The difference between pre-trained and fine-tuned LLM in terms of the model's loss of the member data points (green solid line vs. red solid line in Figure \ref{fig:unlearningMIA}) are usually much larger than that of the non-member data points (green dotted line vs. red dashed line in Figure \ref{fig:unlearningMIA}). Hence, with a predefined threshold, $\epsilon$, samples with a loss-difference higher than $\epsilon$ are considered as belonging to the finetuning dataset.
    Formally, we decide membership based on the rule:
    \begin{align*}
    x \in D_\text{ft}, \quad &\text{if} \quad | \mathcal{L}(\theta_\text{ft}^\text{adv}, x) - \mathcal{L}(\theta_{\text{pre}}, x) | \geq \epsilon, \\
    x \notin D_\text{ft}, \quad &\text{if} \quad | \mathcal{L}(\theta_\text{ft}^\text{adv}, x) - \mathcal{L}(\theta_{\text{pre}}, x) | < \epsilon.
    \end{align*}

\subsection{Data Extraction}

For the data extraction attack, we follow a poisoning algorithm that is very similar to MIA, with some key modifications in the design.

\subsubsection{Poisoning Algorithm for DEA, $T_\text{adv}^\text{de}$:} The attacker creates a noisy version of $D_c$, denoted as $D'_c$ by concatenating each prefix in $P_c$ with some noisy suffixes $S'$, and then runs unlearning on $\theta_\text{pre}$ with this noisy dataset according to equation~\ref{eq:unl_lim}. Just as before, this poisoning approach ensures that the model carries high loss values for these noisy samples before fine-tuning. We utilize two different mechanisms for creating the noisy suffixes:

\ding{114} \textbf{Random word concatenation:} Generate the noisy suffix with a fixed or variable number of random words, which might not have any semantic coherence with each other.

\ding{114} \textbf{Autoregressive generation:} Prompt the pre-trained language model, $\theta_\text{pre}$, with the prefixes to fill out the suffix part. 

After carrying out the poisoning algorithm on the pre-trained LLM, the next few steps of the threat model take place, including model distribution, fine-tuning, and returning the black-box access of the model to the attacker. Finally, the attacker prompts the fine-tuned model with each prefix in $P_c$ and tries to successfully reconstruct the original suffix present in $D_\text{ft}$.

While crafting the noisy samples in DEA based on random word concatenation or autoregressive generation, we add a random number of tokens in a range of 15-20 to the prefix for both cases. Also, we set the default length of known prefixes to 20\% of each full-text sequence. Later, we also do an ablation study by varying the prefix length. Besides, we do ablation with several text generation strategies \cite{gatt2018survey}, including greedy search, beam search decoding, and contrastive search \cite{su2022contrastive}. However, we select beam search decoding with a beam size of 5 as the default configuration for all experiments.

\begin{table*}[t]
\scriptsize
\centering
\begin{tabular}{cl|clccc|clccc}
\hline
\multicolumn{2}{r|}{\textbf{Dataset}} & \multicolumn{5}{c|}{\textbf{MIND}} & \multicolumn{5}{c}{\textbf{Wiki+PII}} \\ \hline
\multicolumn{1}{c|}{\textbf{\begin{tabular}[c]{@{}c@{}}FT \\ Method\end{tabular}}} & \textbf{\begin{tabular}[c]{@{}l@{}}MIA\\ Method\end{tabular}} & \multicolumn{1}{c|}{Val-PPL} & \begin{tabular}[c]{@{}l@{}}Best\\ Acc\end{tabular} & \multicolumn{1}{r}{\begin{tabular}[c]{@{}r@{}}TPR @ \\ 1\%FPR\end{tabular}} & \multicolumn{1}{l}{\begin{tabular}[c]{@{}l@{}}TPR @ \\ 0.1\% FPR\end{tabular}} & AUC & \multicolumn{1}{c|}{Val-PPL} & \begin{tabular}[c]{@{}l@{}}Best\\ Acc\end{tabular} & \multicolumn{1}{l}{\begin{tabular}[c]{@{}l@{}}TPR @ \\ 1\%FPR\end{tabular}} & \multicolumn{1}{l}{\begin{tabular}[c]{@{}l@{}}TPR @ \\ 0.1\% FPR\end{tabular}} & AUC \\ \hline
\multicolumn{1}{c|}{\multirow{8}{*}{\begin{tabular}[c]{@{}c@{}}Full-Ft\\ Llama2-7B\end{tabular}}} & Baseline-loss & \multicolumn{1}{c|}{16.00} & \multicolumn{1}{c}{76.80\%} & 8.20\% & 1.00\% & 79.48\% & \multicolumn{1}{c|}{9.15} & \multicolumn{1}{c}{73.30\%} & 4.80\% & 2.60\% & 77.89\% \\
\multicolumn{1}{c|}{} & Baseline-Rel & \multicolumn{1}{c|}{16.00} & \multicolumn{1}{c}{79.10\%} & 1.60\% & 0.00\% & 81.00\% & \multicolumn{1}{c|}{9.15} & \multicolumn{1}{c}{78.10\%} & 19.20\% & 9.80\% & 84.83\% \\ \cline{2-12} 
\multicolumn{1}{c|}{} & Poison-char-loss & \multicolumn{1}{c|}{16.27} & \multicolumn{1}{c}{81.30\%} & 24\% & \textbf{8.80\%} & 84.72\% & \multicolumn{1}{c|}{11.02} & \multicolumn{1}{c}{83.00\%} & 16.80\% & 5.00\% & 87.88\% \\
\multicolumn{1}{c|}{} & Poison-char-Rel & \multicolumn{1}{c|}{16.27} & \multicolumn{1}{c}{86.40\%} & 2.40\% & 0.40\% & \textbf{88.51\%} & \multicolumn{1}{c|}{11.02} & \multicolumn{1}{c}{\textbf{90.80\%}} & 56.60\% & 32.60\% & 95.60\% \\
\multicolumn{1}{c|}{} & Poison-char-Aux & \multicolumn{1}{c|}{16.02} & \multicolumn{1}{c}{\textbf{87.90\%}} & 21.60\% & 7.60\% & 86.91\% & \multicolumn{1}{c|}{11.15} & \multicolumn{1}{c}{84.60\%} & 23.60\% & 6.40\% & 89.47\% \\ \cline{2-12} 
\multicolumn{1}{c|}{} & Poison-word-loss & \multicolumn{1}{c|}{16.19} & \multicolumn{1}{c}{81.60\%} & 21.40\% & 9\% & 84.83\% & \multicolumn{1}{c|}{11.03} & \multicolumn{1}{c}{83.80\%} & 16.20\% & 5.40\% & 87.89\% \\
\multicolumn{1}{c|}{} & Poison-word-Rel & \multicolumn{1}{c|}{16.19} & \multicolumn{1}{c}{86.50\%} & 2.40\% & 0.00\% & 88.40\% & \multicolumn{1}{c|}{11.03} & \multicolumn{1}{c}{90.40\%} & \textbf{61.60\%} & 30.60\% & \textbf{95.68\%} \\
\multicolumn{1}{c|}{} & Poison-word-Aux & \multicolumn{1}{c|}{16.26} & \multicolumn{1}{c}{82.70\%} & \textbf{23.40\%} & 8.40\% & 86.97\% & \multicolumn{1}{c|}{11.06} & \multicolumn{1}{c}{85.10\%} & 20.20\% & 5.60\% & 89.59\% \\ \hline
\multicolumn{1}{c|}{\multirow{8}{*}{\begin{tabular}[c]{@{}c@{}}Full-Ft\\ GPT-Neo\end{tabular}}} & Baseline-loss & \multicolumn{1}{c|}{64.29} & \multicolumn{1}{c}{70.80\%} & 6.00\% & 2.80\% & 74.53\% & \multicolumn{1}{c|}{19.68} & \multicolumn{1}{c}{71.50\%} & 4.60\% & 1.00\% & 76.58\% \\
\multicolumn{1}{c|}{} & Baseline-Rel & \multicolumn{1}{c|}{64.29} & \multicolumn{1}{c}{79.99\%} & 0.40\% & 0.00\% & 80.70\% & \multicolumn{1}{c|}{19.68} & \multicolumn{1}{c}{84.20\%} & 24.20\% & 14.80\% & 90.64\% \\ \cline{2-12} 
\multicolumn{1}{c|}{} & Poison-char-loss & \multicolumn{1}{c|}{63.44} & \multicolumn{1}{c}{72.30\%} & 9.60\% & 3.60\% & 76.11\% & \multicolumn{1}{c|}{19.75} & \multicolumn{1}{c}{73.40\%} & 10.60\% & 2.80\% & 78.32\% \\
\multicolumn{1}{c|}{} & Poison-char-Rel & \multicolumn{1}{c|}{63.44} & \multicolumn{1}{c}{83.60\%} & 0.60\% & 0.00\% & 86.39\% & \multicolumn{1}{c|}{19.75} & \multicolumn{1}{c}{\textbf{88.90\%}} & \textbf{51.20\%} & 31.00\% & 94.85\% \\
\multicolumn{1}{c|}{} & Poison-char-Aux & \multicolumn{1}{c|}{64.18} & \multicolumn{1}{c}{73.20\%} & \textbf{10.60\%} & 5.20\% & 77.89\% & \multicolumn{1}{c|}{19.74} & \multicolumn{1}{c}{74.40\%} & 25.00\% & 7.00\% & 80.56\% \\ \cline{2-12} 
\multicolumn{1}{c|}{} & Poison-word-loss & \multicolumn{1}{c|}{65.72} & \multicolumn{1}{c}{72.40\%} & 9.60\% & 5.20\% & 76.04\% & \multicolumn{1}{c|}{19.74} & \multicolumn{1}{c}{73.50\%} & 10.20\% & 3.00\% & 78.36\% \\
\multicolumn{1}{c|}{} & Poison-word-Rel & \multicolumn{1}{c|}{65.72} & \multicolumn{1}{c}{\textbf{83.90\%}} & 0.60\% & 0.00\% & 86.36\% & \multicolumn{1}{c|}{19.74} & \multicolumn{1}{c}{88.10\%} & 51.00\% & 32.60\% & \textbf{94.89\%} \\
\multicolumn{1}{c|}{} & Poison-word-Aux & \multicolumn{1}{c|}{66.72} & \multicolumn{1}{c}{73.20\%} & 10.20\% & 5.40\% & \textbf{77.77\%} & \multicolumn{1}{c|}{19.75} & \multicolumn{1}{c}{73.40\%} & 25.20\% & 7.00\% & 80.60\% \\ \hline
\multicolumn{1}{c|}{\multirow{8}{*}{\begin{tabular}[c]{@{}c@{}}LoRA-Ft\\ Llama2-7B\end{tabular}}} & Baseline-loss & \multicolumn{1}{c|}{17.04} & 63.10\% & 5.20\% & 0.20\% & 67.32\% & \multicolumn{1}{c|}{9.14} & 60.00\% & 3.20\% & 0.20\% & 62.66\% \\
\multicolumn{1}{c|}{} & Baseline-Rel & \multicolumn{1}{c|}{17.04} & 71.10\% & 0.00\% & 0.00\% & 74.62\% & \multicolumn{1}{c|}{9.14} & 65.30\% & 7.40\% & 1.00\% & 69.24\% \\ \cline{2-12} 
\multicolumn{1}{c|}{} & Poison-char-loss & \multicolumn{1}{c|}{16.64} & 66.60\% & 6.40\% & 3.20\% & 69.25\% & \multicolumn{1}{c|}{9.17} & 61.40\% & 3.20\% & 0.40\% & 63.32\% \\
\multicolumn{1}{c|}{} & Poison-char-Rel & \multicolumn{1}{c|}{16.64} & 76.50\% & 0.20\% & 0.30\% & \textbf{81.00\%} & \multicolumn{1}{c|}{\textbf{9.17}} & \textbf{72.00\%} & 7.80\% & 4.40\% & \textbf{76.70\%} \\
\multicolumn{1}{c|}{} & Poison-char-Aux & \multicolumn{1}{c|}{17.55} & 64.50\% & 6.20\% & 2.40\% & 67.77\% & \multicolumn{1}{c|}{8.94} & 60.50\% & 10\% & 4.80\% & 63.63\% \\ \cline{2-12} 
\multicolumn{1}{c|}{} & Poison-word-loss & \multicolumn{1}{c|}{16.77} & 66.50\% & 4.80\% & 2.60\% & 69.39\% & \multicolumn{1}{c|}{9.13} & 60.80\% & 2.00\% & 0.10\% & 62.44\% \\
\multicolumn{1}{c|}{} & Poison-word-Rel & \multicolumn{1}{c|}{16.77} & 77.90\% & 0.60\% & 0.40\% & \textbf{81.05\%} & \multicolumn{1}{c|}{\textbf{9.13}} & \textbf{71.50\%} & 10.60\% & 1.50\% & 75.80\% \\
\multicolumn{1}{c|}{} & Poison-word-Aux & \multicolumn{1}{c|}{16.67} & 64.50\% & 6.20\% & 2.00\% & 67.92\% & \multicolumn{1}{c|}{9.00} & 61.90\% & 10.00\% & 5.60\% & 65.46\% \\ \hline
\multicolumn{1}{c|}{\multirow{8}{*}{\begin{tabular}[c]{@{}c@{}}QLoRA-Ft\\ (4 bit)\\ Llama2-7B\end{tabular}}} & Baseline-loss & \multicolumn{1}{c|}{17.35} & 63.70\% & 5.20\% & 1.00\% & 67.60\% & \multicolumn{1}{c|}{9.07} & 59.90\% & 2.80\% & 0.20\% & 61.96\% \\
\multicolumn{1}{c|}{} & Baseline-Rel & \multicolumn{1}{c|}{17.35} & 71.40\% & 0.20\% & 0.00\% & 74.70\% & \multicolumn{1}{c|}{9.07} & 65.10\% & 6.00\% & 1.00\% & 69.02\% \\ \cline{2-12} 
\multicolumn{1}{c|}{} & Poison-char-loss & \multicolumn{1}{c|}{17.42} & 65.00\% & 6.60\% & 3.40\% & 67.47\% & \multicolumn{1}{c|}{9.28} & 61.20\% & 3.60\% & 0.80\% & 62.27\% \\
\multicolumn{1}{c|}{} & Poison-char-Rel & \multicolumn{1}{c|}{17.42} & 76.70\% & 0.20\% & 0.00\% & 79.02\% & \multicolumn{1}{c|}{9.28} & 70.70\% & 7.00\% & 2.80\% & 75.66\% \\
\multicolumn{1}{c|}{} & Poison-char-Aux & \multicolumn{1}{c|}{16.75} & 66.30\% & 7.40\% & 2.80\% & 69.37\% & \multicolumn{1}{c|}{9.17} & 61.10\% & 10\% & 3.80\% & 63.84\% \\ \cline{2-12} 
\multicolumn{1}{c|}{} & Poison-word-loss & \multicolumn{1}{c|}{17.22} & 64.60\% & 6.80\% & 3.20\% & 67.20\% & \multicolumn{1}{c|}{9.28} & 61.00\% & 2.60\% & 0.40\% & 62.67\% \\
\multicolumn{1}{c|}{} & Poison-word-Rel & \multicolumn{1}{c|}{17.22} & 77.00\% & 0.40\% & 0.00\% & \textbf{80.12\%} & \multicolumn{1}{c|}{\textbf{9.28}} & \textbf{71.30\%} & 9.60\% & 3.00\% & \textbf{75.81\%} \\
\multicolumn{1}{c|}{} & Poison-word-Aux & \multicolumn{1}{c|}{16.79} & 66.00\% & 7.00\% & 3.20\% & 69.67\% & \multicolumn{1}{c|}{9.26} & 61.90\% & 10.40\% & 5.00\% & 65.55\% \\ \hline
\end{tabular}
\caption{ Membership inference evaluation with different finetuning methods.}
\label{tab: mia_all}
\end{table*}

\section{Experimental Setup}

In this section, we discuss the default settings and hyperparameters used for different experiments. 
\subsection{Dataset} We perform experiments on two datasets, each representing a particular data type. The first dataset consists of news article abstracts obtained from a subset of the Microsoft News Dataset (MIND) \cite{wu2020mind}. It has three partitions: train, test, and validation. We took a subset of 20K training samples for fine-tuning, 1K subset of validation samples, and 1K test samples. We selected this dataset to investigate how our attacks perform to leak the privacy of general-purpose English texts from the fine-tuning dataset. The second dataset is a fusion of Wikitext-103 \cite{Merity2017RegularizingAO} and 
AI4Privacy\footnote{https://huggingface.co/datasets/ai4privacy/pii-masking-200k}. The latter is an open-source
privacy dataset that holds real-life personal identifiable information (PII) data points. We inject 1,000 randomly selected samples from AI4Privacy into the WikiText-103 dataset. This dataset is meant to experiment with how our attacks are able to extract private information such as addresses, phone
numbers, passwords, etc.

\subsection{Models and Fine-Tuning Methods}

To evaluate our attacks we select two different families of large language models, GPT-Neo 1.3 billion parameter variant\footnote{https://huggingface.co/EleutherAI/gpt-neo-1.3B} from EleutherAI and Llama-2 7 billion parameter variant\footnote{https://huggingface.co/meta-llama/Llama-2-7b} from Meta. 
Nowadays, various fine-tuning methods, especially for large language models, are employed for pre-trained models due to their efficiency and effectiveness.
Since an adversary may not have control over the
fine-tuning algorithm, we demonstrate how effective our attacks are against different fine-tuning methods. We trained the Llama-2 model using full fine-tuning (Full-FT), LoRA-FT \cite{hu2021LoRA}, and 4-bit QLoRA \cite{dettmers2024qLoRA}. We set a default learning rates for Full-FT, LoRA-FT, and QLoRA-FT as $2\times10^{-5}$, $2\times10^-{4}$, and $2\times10^{-4}$, respectively, and trained for 5 epochs with early stopping to prevent overfitting. 

\subsection{Evaluation Metrics}
We use the perplexity on the validation dataset(Val-PPL$\downarrow$) to measure the utility of the fine-tuned model, as well as the stealthiness of our proposed attacks. \cite{carlini2022membership} pioneered the practice of analyzing True Positive Rate (TPR$\uparrow$) at low False Positive Rate (FPR) thresholds to highlight the effectiveness of attacks under stringent conditions. Following this approach, our evaluation framework employs several key metrics: TPR at 0.01\% FPR, TPR at 0.1\% FPR, Area Under the Curve (AUC$\uparrow$), and Best Accuracy (Best Acc$\uparrow$), defined as the maximum accuracy achieved along the tradeoff curve.
On the other hand, to evaluate data extraction, we compute the number of successful reconstructions (NSR$\uparrow$), i.e., the number of extracted sequences that are part of the finetuning dataset. 

\section{Results}
In this section, we provide a comprehensive evaluation of our proposed attacks and discuss the experimental outcomes from various critical perspectives.

\subsection{Membership Inference}
To evaluate the membership inference attack (MIA), we take 1K test sequences, 500 of which are member samples, i.e., present in the fine-tuning dataset, and the remaining 500 are non-member samples, i.e., absent in the fine-tuning dataset. 
\subsubsection{Baselines and Proposed Attacks:}
We consider two baseline MIA: the first one is simply based on model loss (Baseline-Loss), with the assumption that member data points would have a lower loss value than the non-member samples. The second baseline is based on relative loss with respect to the pre-trained model (Baseline-Rel), i.e., the loss difference between fine-tuned and the pre-trained models, where the relative loss of member samples should be higher than the non-member samples. Apart from that, as mentioned in the Methodology section, for both character perturbation and word perturbation-based poisoning, we adopt three inference strategies- simple loss-based (Poison-char/word-Loss), reference data-based (Poison-char/word-Aux) and reference model-based (Poison-char/word-Rel). 

\subsubsection{Model Utility/ Stealthiness:}
Table \ref{tab: mia_all} compares the attack performance and model utility of Llama2-7B on two datasets, MIND and Wiki-PII, with respect to different MIA configurations for full fine-tuning, LoRA and QLoRA finetuning. It also contains the results for GPT-Neo with Full-Ft. If we compare the poisoning methods with the baselines (i.e., no poisoning), one important observation is that the change in validation perplexity after incorporating the poisoning is negligible for both the Llama2 and GPT-Neo models and across different fine-tuning algorithms. This indicates that our poisoning methods are stealthy enough to surpass all the detection measures based on model loss. Besides, Llama2-7B generally has lower Val-PPL on both datasets compared to GPT-Neo, indicating its better generalization ability. 

\subsubsection{Attack Performance:}
In a nutshell, our proposed MIA methods significantly outperform the two baselines for both datasets with respect to all evaluation metrics for full finetuning (Table \ref{tab: mia_all}). Firstly, if we consider MIA for general-purpose English texts, i.e., the \textbf{MIND} dataset on the Llama2 model, the reference model-based attacks (Poison-char-Rel and Poison-word-Rel) improve the AUC by $\sim$7.5\% and the Best Acc by $\sim$7\% over baseline. Additionally, the reference data-based attacks (Poison-char-Aux and Poison-word-Aux) show superior performance in the low-FPR region, improving the TPR at 1\% FPR by 15-20\% compared to the baseline.

On the other hand, looking at the MIA results on Llama2 for PII texts, i.e., the \textbf{Wiki+AI4Privacy} dataset, we can find even more promising results. The reference model-based attacks derive nearly 96\% AUC and $\sim$91\% Best Acc score, beating the two baselines by 11-18\% and 12-17\% respectively. Unlike the MIND dataset, here, reference model-based attacks perform better than reference data-based attacks in the low-FPR region, as Poison-word-Rel begets an attractive TPR of $\sim$62\% at 1\% FPR and Poison-char-Rel gives $\sim$33\% TPR at 0.1\% FPR. 

In summary, the Llama2 model is more vulnerable to our proposed MIA attacks on PII data than plain English texts. In addition to that,  reference data-based attacks demonstrate better performance for plain English texts, while reference model-based attacks perform better for PII data.

Moreover, if we take a look at the results for the GPT-Neo model in Table \ref{tab: mia_all} we will find a similar improvement in attack performance over the baselines. However, the scores (AUC, Best Acc, TPR at low-FPR region) are overall lower for GPT-Neo compared to Llama2. One possible reason for this is the size of the language model. Prior work \cite{carlini2022quantifying} has also shown that larger LMs memorize more than the smaller ones. 

\subsubsection{Ablation Studies:\\}

I) \textbf{Finetuning methods:} By comparing the results among different finetuning methods in Table \ref{tab: mia_all}, we can deduce that both of these parameter-efficient finetuning methods such as LoRA and QLoRA, have been effective in reducing the success rate of membership inference attacks without significantly impacting the model's utility.  LoRA finetuning, in particular, resulted in a lower validation perplexity than full fine-tuning on the wiki+PII dataset. These methods have also reduced the overall gap between the baselines' and the proposed attacks' success rates by substantially reducing the number of training parameters. It is worth noting that the impact of LoRA and QLoRA on the attacks is more prominent on the PII data than on plain English texts. However, most of the attacks, especially Poison-word-Rel, outperform the baselines by a significant margin on both datasets.\\
II) \textbf{Noising-level:} 
\begin{figure}[t]
\centering
\includegraphics[width=0.8\linewidth]{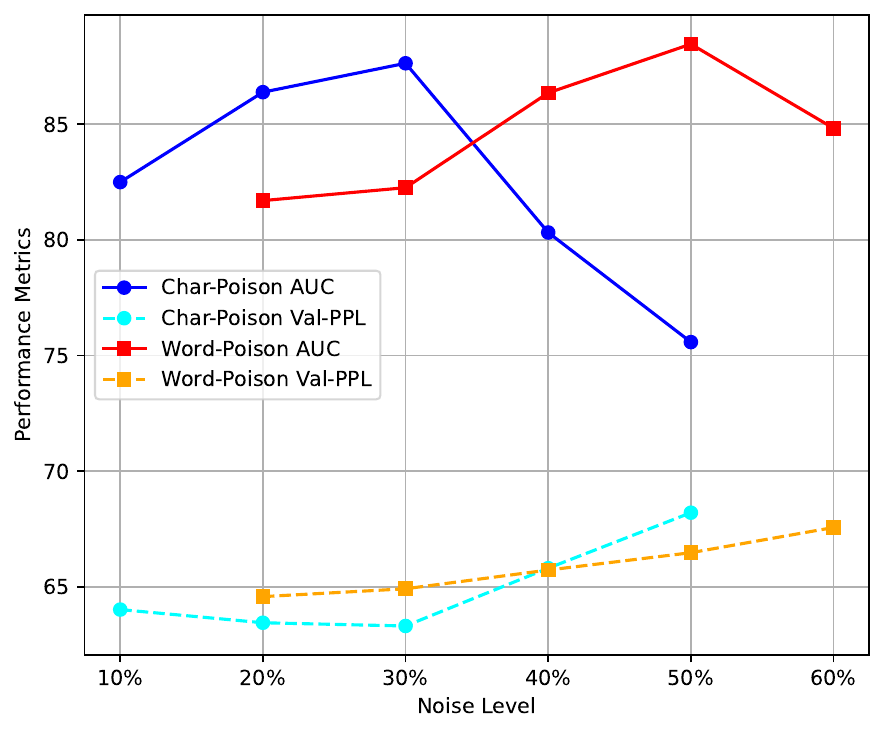}
\caption{Membership inference AUC and validation perplexity for random Poison-char-Rel and Poision-word-Rel attacks with varying noising level.}
\label{fig:ablation_with_noise}
\end{figure}
Figure \ref{fig:ablation_with_noise} provides a comparison between Char-Poison-Rel and Word-Poison-Rel methods under varying noise levels. The Char-Poison method shows an optimal attack performance at a 30\% noise level, but its effectiveness decreases as noise increases further. This is because, when the noise is too heavy, the noisy samples lose coherence with their original counterparts, hence deviating from the goal of the proposed method. On the other hand, Word-Poison proves more resilient, improving attack efficacy up to a 50\% noise level. However, this comes at the cost of a higher increase in Validation Perplexity, indicating a more substantial degradation in model utility as noise levels rise. One interesting finding is the reduction of Val-PPL up to 30\% noise level, which indicates that unlearning on noisy data can potentially enhance the model utility to some extent. 

\subsection{Benchmark Study}
\begin{figure}[t]
\centering
\includegraphics[width=0.9\linewidth, trim={0 7 0 7},clip]{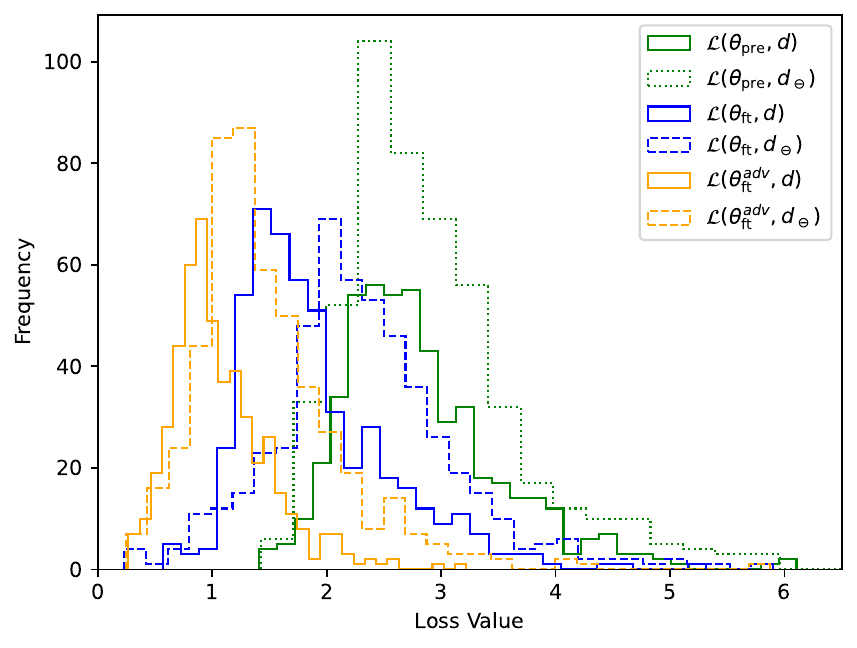}
\caption{Histograms of loss values on pre-trained model $\theta_\text{pre}$, fine-tuned model $\theta_\text{ft}$, and fine-tuned poisoned (\cite{wen2024privacy}) model $\theta_\text{ft}^\text{adv}$.}
\label{fig: prelearningMIA}
\end{figure}

\begin{table}[t]
\scriptsize
\begin{tabular}{l|l|lll|lll}
\hline
\textbf{\begin{tabular}[c]{@{}l@{}}Ft \\ Method\end{tabular}} & \textbf{Dataset} & \multicolumn{3}{c|}{\textbf{MIND}} & \multicolumn{3}{c}{\textbf{Wiki+PII}} \\ \hline
 & \textbf{Model} & \begin{tabular}[c]{@{}l@{}}Base-\\ line\end{tabular} & \begin{tabular}[c]{@{}l@{}}DEA\\ Gen\end{tabular} & \begin{tabular}[c]{@{}l@{}}DEA\\ Rand\end{tabular} & \begin{tabular}[c]{@{}l@{}}Base-\\ line\end{tabular} & \begin{tabular}[c]{@{}l@{}}DEA\\ Gen\end{tabular} & \begin{tabular}[c]{@{}l@{}}DEA\\ Rand\end{tabular} \\ \hline
\multirow{2}{*}{Full} & Llama2 & 93 & 177 & 124 & 8 & 32 & 15 \\
 & GPT-Neo & 79 & 120 & 91 & 42 & 103 & 68 \\ \hline
LoRA & Llama2 & 6 & 18 & 10 & 0 & 5 & 0 \\ \hline
QLoRA & Llama2 & 5 & 17 & 10 & 0 & 0 & 0 \\ \hline
\end{tabular}
\caption{Data extraction attack evaluation for two LLMs, two benchmark datasets, and four different fine-tuning methods. NSR (Number of Successful Reconstruction) is calculated out of 500 test samples for each dataset.}
\label{tab:dea_all}
\end{table}

\begin{table*}[t]
\centering
\scriptsize
\begin{tabular}{l|cccc|cccc|cccc}
\hline
\multicolumn{1}{r|}{\textbf{$\epsilon=$}} & \multicolumn{4}{c|}{\textbf{10}} & \multicolumn{4}{c|}{\textbf{50}} & \multicolumn{4}{c}{\textbf{$\infty$}} \\ \hline
\textbf{Attack} & \multicolumn{1}{l}{Val-PPL} & \multicolumn{1}{l}{\begin{tabular}[c]{@{}l@{}}TPR @ \\ 1\% FPR\end{tabular}} & \multicolumn{1}{l}{AUC} & \multicolumn{1}{l|}{NSR} & \multicolumn{1}{l}{Val-PPL} & \multicolumn{1}{l}{\begin{tabular}[c]{@{}l@{}}TPR @ \\ 1\% FPR\end{tabular}} & \multicolumn{1}{l}{AUC} & \multicolumn{1}{l|}{NSR} & \multicolumn{1}{l}{Val-PPL} & \multicolumn{1}{l}{\begin{tabular}[c]{@{}l@{}}TPR @ \\ 1\% FPR\end{tabular}} & \multicolumn{1}{l}{AUC} & \multicolumn{1}{l}{NSR} \\ \hline
MIA-Baseline-loss & 101.07 & 2.60\% & 50.62\% & - & 96.80 & 3.00\% & 51.61\% & - & 67.53 & 5.60\% & 68.18\% & - \\
MIA-Poison-char-Rel & 101.98 & 1.40\% & 61.20\% & - & 96.85 & 1.80\% & 64.02\% & - & 66.63 & 2.20\% & 86.18\% & - \\
MIA-Poison-char-Aux & 100.87 & 3.20\% & 53.32 & - & 96.50 & 3.40\% & 54.52\% & - & 71.03 & 14.60\% & 75.23\% & - \\ \hline
DEA-Baseline & 101.07 & - & - & 0 & 96.80 & - & - & 0 & 67.53 & - & - & 8 \\
DEA-Gen & 100.48 & - & - & 4 & 96.88 & - & - & 5 & 65.11 & - & - & 19 \\ \hline
\end{tabular}
\caption{Membership inference and data extraction results with differential privacy defense.}
\label{tab:dp_ft}
\end{table*}
We simulated the concurrent work of \citet{wen2024privacy} by minimizing the loss of the target data points ($D_c$) on the pre-trained Llama2-7B model to get poisoned model. As we mentioned in the Related Work section before, their approach tends to overfit both the member and non-member data samples of $D_c$. The empirical studies further verify this. More specifically, Figure \ref{fig: prelearningMIA} shows the histograms of loss values of member data $d$ and non-member data $d_\ominus$, on pre-trained model, $\theta_{pre}$ (green color) and fine-tuned model, $\theta_{ft}$ (blue color). It also shows the histograms of loss values of $d$ and $d_\ominus$ after fine-tuning on the poisoned (via loss minimization strategy of \citet{wen2024privacy}) model, $\theta_{ft}^{adv}$ (orange color). In both cases (orange and red bars in Figure \ref{fig: prelearningMIA} and \ref{fig:unlearningMIA} respectively), the model's loss significantly drops after finetuning. However, after finetuning on the poisoned model by \citet{wen2024privacy}, the loss difference between $d$ and $d_\ominus$ (difference between orange solid line and orange dashed line in Figure \ref{fig: prelearningMIA}) is small, and much smaller than that of finetuning on the proposed poisoned model (difference between red solid line and red dashed line in Figure \ref{fig:unlearningMIA}).  

Thus, we empirically found that the membership inference performance by their method is just comparable to the baseline performance on our tested dataset, whereas our attacks substantially outperform the baselines in almost all possible setups. For the membership inference attack by \citet{wen2024privacy}, it achieves Best Acc= 71.5\%, FPR= 9.6\% when TPR@1\%, FPR= 2.2\% when TPR@ 0.1\%, and AUC= 76.63\% on Llama2-7B model and MIND dataset. Comparing with the corresponding results in Table 1 in the main body of the paper, we can see that our proposed methods are significantly better.

\subsection{Data Extraction}

To evaluate the data extraction attack (DEA), we take 500 test sequences (PII sequences in the case of Wiki+AI4Privacy) from the training dataset.
\subsubsection{Baseline and Proposed Attacks:} We adopt a simple baseline similar to \citet{carlini2019secret, carlini2021extracting} where we prompt the fine-tuned LLM with the known prefixes and get the highest likelihood generated sequences. Besides, as mentioned in the Methodology section, we propose two poisoning methods for data extraction- random word concatenation (DEA-Rand) and autoregressive generation (DEA-Gen). 

\subsubsection{Attack Performance:} Table \ref{tab:dea_all} demonstrates the data extraction results in terms of NSR (number of successful reconstructions) against Llama2-7B and GPT-Neo 1.3B models for two datasets and three different finetuning methods. In the case of full fine-tuning, our autoregressive generation-based attack method (DEA-Gen) derives attractive NSR against both Llama2 and GPT-Neo. However, the DEA-Rand attack, while surpassing the baseline performance, did not perform as well as the DEA-Gen. Interestingly, Llama2 showed more resilience against DEA attacks on personally identifiable information (PII) data than on plain English texts. Additionally, similar to the MIA results for LoRA and QLoRA finetuning, these two methods have also shown greater robustness against data extraction attacks for both language models and the datasets.
\subsubsection{Ablation Studies:\\}
I) \textbf{Prefix length:} Table \ref{tab:de_ablation} shows the NSR scores for varying lengths (denoted as the fraction/percentage of each full-text sequence) of known prefixes through which the attacker prompts the model. Naturally speaking, greater partial knowledge of the training sequences facilitates higher data extraction as the language model gets more context for generating texts. Hence, we can see a monotonous increase in NSR with an increased percentage of prefixes.\\
II) \textbf{Sequence Repetition:} It happens quite often in real-world datasets that some sequences occur multiple times. Previous studies \cite{lee2021deduplicating, carlini2022quantifying} have indicated that duplicate sequences in the training set can lead to increased memorization in LLMs. Our experimental results in Table \ref{tab:de_ablation} support this finding. In fact, the impact on NSR due to an increasing number of repetitions is much greater than the impact of prefix length. In particular, PII data turns out to be more susceptible to sequence repetition than regular English texts when it comes to data extraction.\\
\begin{table}[]
\scriptsize
\centering
\begin{tabular}{l|lllll}
\hline
\textbf{Prefix length} & \textbf{10\%} & \textbf{20\%} & \textbf{30\%} & \textbf{40\%} & \textbf{50\%} \\ \hline
MIND-NSR     & 95  & 177 & 208 & 259 & 326 \\
Wiki+PII-NSR & 11  & 32  & 51  & 57  & 72  \\ \hline
\textbf{Repetition}    & \textbf{1}    & \textbf{3}    & \textbf{5}    & \textbf{10}   & \textbf{15}   \\ \hline
MIND-NSR     & 177 & 268 & 349 & 457 & 466 \\
Wiki-PII-NSR & 32  & 107 & 245 & 402 & 430 \\ \hline
\end{tabular}
\caption{Ablation studies on data extraction attacks for varying prefix length and sequence repetition.}
\label{tab:de_ablation}
\end{table}
II) \textbf{Text Generation Methods:}
\begin{table*}[t]
\scriptsize
\centering
\begin{tabular}{l|l|c|ccc|cccc}
\hline
\textbf{Dataset} & \textbf{DEA Method} & \textbf{Greedy} & \textbf{Beam-3} & \textbf{Beam-5} & \textbf{Beam-7} & \textbf{\begin{tabular}[c]{@{}c@{}}Contrastive \\ alpha=0.5 top-k=5\end{tabular}} & \textbf{\begin{tabular}[c]{@{}c@{}}Contrastive \\ alpha=0.5 top-k=7\end{tabular}} & \textbf{\begin{tabular}[c]{@{}c@{}}Contrastive \\ alpha=0.3 top-k=3\end{tabular}} & \textbf{\begin{tabular}[c]{@{}c@{}}Contrastive \\ alpha=0.1 top-k=3\end{tabular}} \\ \hline
\multirow{2}{*}{MIND} & Baseline & 46 & 73 & 93 & 105 & 32 & 34 & 32 & 42 \\
 & DEA-Gen & 111 & 155 & 177 & 187 & 95 & 92 & 93 & 96 \\ \hline
\multirow{2}{*}{Wiki+PII} & Baseline & 11 & 15 & 8 & 10 & 3 & 2 & 5 & 2 \\
 & DEA-Gen & 29 & 28 & 32 & 34 & 17 & 23 & 27 & 19 \\ \hline
\end{tabular}
\caption{Data extraction results in terms of NSR (Number of Successful Reconstruction) on Llama2-7B model for different text generation methods. NSR is calculated out of 500 test samples for each dataset.}
\label{tab:dea_gen_ablation}
\end{table*}
Table \ref{tab:dea_gen_ablation} presents an ablation study evaluating the data extraction attack for various text generation methods on the NSR (Number of Successful Reconstructions) metric, applied to the Llama2-7B model across two datasets: MIND and Wiki+AI4Privacy. The methods compared include Greedy, Beam Search with different beam widths (3, 5, 7), and Contrastive Search with varying configurations (penalty alpha and top-k).
Here, Beam Search consistently outperforms Greedy decoding across both datasets, with the NSR improving as the beam width increases. For instance, on the MIND dataset, the NSR rises from 46 with Greedy, to 105 (out of 500 samples) with Beam-7 in the Baseline method, showing a clear advantage of using a wider beam for data extraction. However, Contrastive Search shows variation in performance depending on the alpha and top-k configurations. Notably, while the Baseline method results in lower NSR values (e.g., 32-42 on MIND), the DEA-Gen method consistently achieves higher NSR (e.g., 92-96 on MIND), especially when tuning the alpha and top-k parameters. The best performance is observed with an alpha of 0.1 and top-k of 3. Apart from that, the MIND dataset, as usual, exhibits higher NSR values compared to PII, indicating that the characteristics of the dataset play a role in the effectiveness of different text generation methods.

\subsection{Effectiveness under Defense}
We adopt differential privacy (DP) \cite{yu2021differentially, li2021large}, a standard defense mechanism in machine learning privacy, and we use the $(\epsilon, \delta)$ implementation of DP-transformers \cite{dp-transformers}. In Table \ref{tab:dp_ft}, we present the effectiveness of our proposed MIA and DEA attacks , as well as the impact on model utility with increasing privacy budget in DP. Overall, under stringent DP finetuning, our proposed MIA attacks achieve a better AUC and slightly worse TPR (except for Poison-Char-Aux) at the lower FPR region. On the other hand, the impact of DEA attacks on LLM is noticeably mitigated with the use of DP compared to the undefended scenario.  However, even with a very relaxed privacy budget (e.g., $\epsilon \leq 50$), applying DP significantly decreases model utility, making the model almost unusable. Thus, the trade-off between utility and privacy raises doubts about the effectiveness of this defense mechanism.

\section{Conclusion}

We developed a novel unlearning-based model poisoning method that amplifies privacy breaches during fine-tuning. Extensive empirical studies demonstrate the proposed method's efficacy on both membership inference and data extraction attacks. Given that the attack is stealthy enough to bypass detection-based defenses and that differential privacy cannot effectively defend against the attacks without significantly impacting model utility, it is important to explore more effective defenses for such poisoning attacks in the future.

\bibliography{aaai25}

\end{document}